\documentclass[conference,a4paper]{APSIPA2025}
\usepackage{amsmath}
\usepackage{graphicx}
\usepackage{multirow}
\usepackage{threeparttable}
\usepackage[backend=biber,style=ieee,]{biblatex}
\usepackage{adjustbox}
\usepackage{float} 
\usepackage[table,xcdraw]{xcolor}
\usepackage{tabularx}
\usepackage{hyperref}
\usepackage{color}
\usepackage{booktabs} 

\usepackage{geometry}
\usepackage{fancyhdr}

\fancypagestyle{firststyle}{
  \fancyhf{}
  \fancyhead[C]{2025 Asia Pacific Signal and Information Processing Association Annual Summit and Conference (APSIPA ASC)}
}

\begin{document}

\title{Dialect Identification Using Resource-Efficient Fine-Tuning Approaches}

\author{
\authorblockN{
Zirui Lin\authorrefmark{1},
Haris Gulzar\authorrefmark{2},
Monnika Roslianna Busto\authorrefmark{2},
Akiko Masaki\authorrefmark{2},
Takeharu Eda\authorrefmark{2} and
Kazuhiro Nakadai\authorrefmark{1}
}

\authorblockA{
\authorrefmark{1}
Dept. of Systems and Control Engineering, School of Engineering, Institute of Science Tokyo, Tokyo, Japan \\
Email: \{linzirui, nakadai\}@ra.sc.e.titech.ac.jp}

\authorblockA{
\authorrefmark{2}
NTT Software Innovation Center, Tokyo, Japan \\
Email: \{haris.gulzar, monikka.busto, takeharu.eda, akiko.masaki\}@ntt.com}
}

\maketitle
\thispagestyle{firststyle}
\pagestyle{fancy}

\begin{abstract}
Dialect Identification (DI) is a task to recognize different dialects within the same language from a speech signal. DI can help to improve the downstream speech related tasks even when speakers have a strong dialect. However, fine-tuning a speech model for tasks like DI is expensive in terms of computation cost and memory requirement. Recent studies have explored fine-tuning pre-trained speech models for tasks like DI using Parameter-Efficient Fine-Tuning (PEFT) methods, which offer parameter efficiency but limited improvement in memory efficiency and training speed. To address these challenges, we explore Memory-Efficient Fine-Tuning (MEFT) methods, originally proposed for language processing, and apply them to the general-purpose pre-trained speech model. We then comprehensively analyze the GPU memory usage and fine-tuning speed based on various MEFT methods.  As a case study, we fine-tune the Whisper model to identify six Mandarin subdialects from the KeSpeech dataset, reducing GPU memory usage by up to 73.25\% and accelerating training speed by a factor of 2.1, while maintaining accuracy comparable to vanilla fine-tuning and PEFT methods.
\end{abstract}

\section{Introduction}
Dialect identification (DI)~\cite{DI-1, DI-2-LID} amounts to identifying dialects belonging to the same language branch, which is a specific case of language identification~\cite{DI-2-LID, LID} (LID) task. However, DI is more challenging than LID because dialects share similar acoustic and linguistic characteristics compared to different languages~\cite{DI-survey, DI-gaussian}.
Dialect identification can help improve speech recognition systems and is also expected to enhance human-computer interaction applications and secure remote access communications~\cite{DI-Contribution}. Moreover, DI can help drive novel e-health and telemedicine services, which can be especially valuable for elderly and homebound individuals.~\cite{DI-survey}

Fine-tuning a pre-trained speech model for downstream tasks has been proved to achieve outstanding performance on various tasks as it can leverage the common speech features learned from large corpus during the pre-training~\cite{FT-for-tasks, adapter4, wav2vec2, xlsr, xls-r}.
However, vanilla fine-tuning, which updates patameters of the whole model, demands substantial computational power and memory capacity, making it a costly endeavor, especially in the case that the space complexity of transformer~\cite{transformer} is $O(n^2)$ while $n$ is the sequence length~\cite{O2}, and the sequence lengths of speech models are often longer than that of language models. To alleviate these resource requirements, recent studies~\cite{DI-PEFT-1, DI-PEFT-2} have adopted Parameter-Efficient Fine-Tuning (PEFT) methods~\cite{peft}, which update only a small subset of model parameters. PEFT drastically reduces compute and storage overhead while achieving performance on par with full fine-tuning and mitigating catastrophic forgetting~\cite{catastrophioc-forgetting}.

Although Parameter-Efficient Fine-Tuning (PEFT) methods significantly reduce the number of parameters to be updated, their gains in memory efficiency are modest, typically ranging from 10\% to 30\%. In other words, PEFT methods provide limited advantages for training on low-resource GPUs, as they still require backpropagation through the entire backbone model to compute gradients for parameter updates~\cite{LST}. This necessitates storing intermediate activations from the backbone model's layers in GPU memory, leading to substantial memory consumption. Furthermore, our observations suggest that PEFT methods have a limited impact on accelerating the training, with improvements of up to about 10\% in our evaluations, which offers only marginal benefits for low-resource GPUs. 

To better adapt fine-tuning for DI tasks on low-resource GPUs, we introduce Memory-Efficient Fine-Tuning (MEFT) methods~\cite{LST, UniPT, SHERL} by using DI as an example. MEFT methods have achieved success in Natural Language Processing (NLP) tasks~\cite{LST, UniPT, SHERL}, but its effectiveness on speech related tasks remains to be explored. We deploy MEFT methods on the general-purpose speech recognition architecture, Whisper~\cite{Whisper} from OpenAI, which has been pre-trained on a large speech corpus, and fine-tune it for DI tasks. MEFT methods construct a lightweight side network alongside the backbone model, taking the intermediate activations of the backbone model as inputs and updating only the side network. This approach avoids backpropagation through the backbone model, reducing GPU memory usage. 

In this work, we evaluate our approaches on subdialects of Mandarin Chinese, as the differences between Mandarin subdialects are relatively subtle compared to dialect-level variations and difficult to be effectively captured by the model~\cite{Chinese-survey}, making their distinction more challenging. Mutual intelligibility is generally high among Mandarin subdialects despite regional lexical differences. This serves as a rigorous test of the capability.

The main contributions of this work are as follows:
\begin{enumerate}
    \item We extend the application of MEFT methods to speech processing, systematically addressing the challenges posed by long input sequences and the computational overhead of side-layer operations. Our analysis demonstrates the effectiveness of MEFT in mitigating GPU memory bottlenecks in fine-tuning speech models. 
    
    \item We report substantial performance gains on the KeSpeech dataset \cite{kespeech}, specifically targeting six Mandarin sub-dialects, where our methods consistently outperform baseline systems in subdialect identification accuracy. 
    
    \item Our proposed approaches achieve comparable accuracy to full fine-tuning (within 0.67\%) while reducing GPU memory consumption by up to 3.7× and improving training throughput by up to 2.1×.  Our code is available at \textit{https://github.com/linzr25/Whisper-DI-MEFT}.
\end{enumerate}

\begin{figure*}[t]
    \centering
    \includegraphics[width=\linewidth]{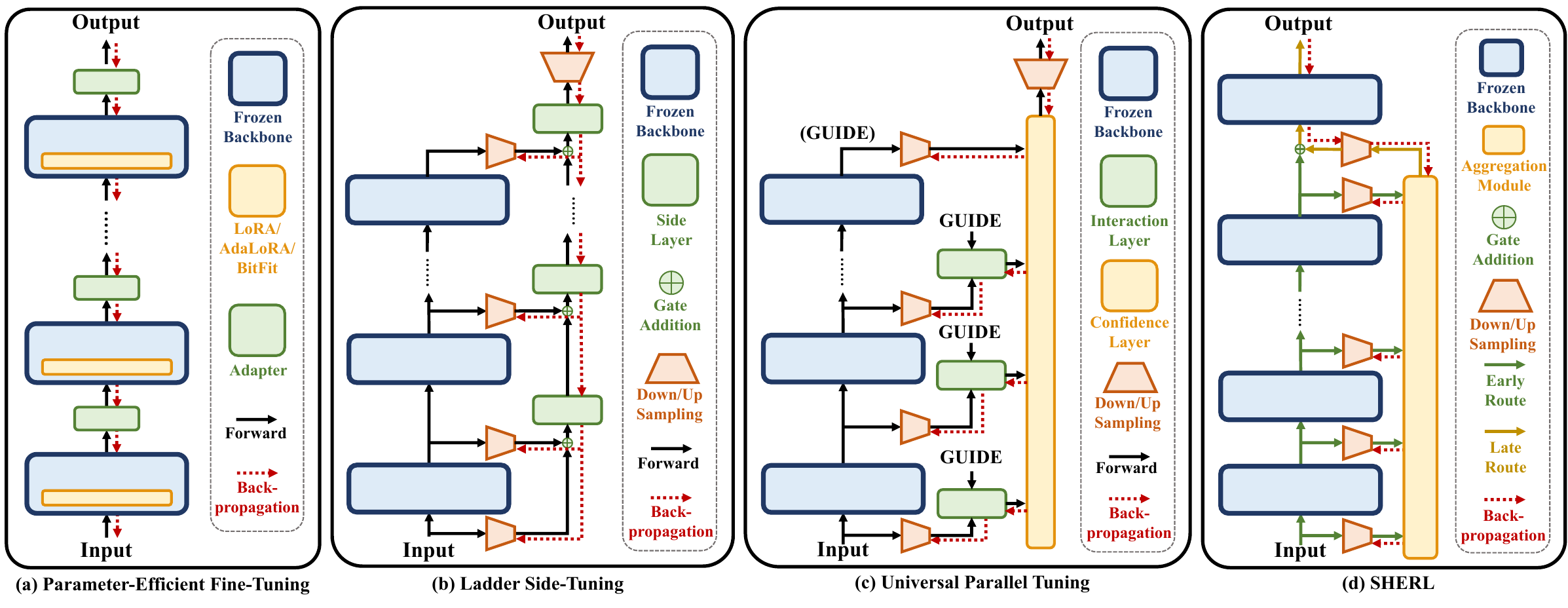}
    \caption{\textbf{(a)} parameter-efficient fine-tuning, including Adapter~\cite{adapter1}, LoRA~\cite{LoRA}, AdaLoRA~\cite{AdaLoRA} and BitFit~\cite{bitfit}; \textbf{(b)} Ladder Side-Tuning (LST)~\cite{LST}; \textbf{(c)} Universal Parallel Tuning (UniPT)~\cite{UniPT}; \textbf{(d)} SHERL~\cite{SHERL}.}
    \label{fig:PEFT_MEFT}
\end{figure*}

\section{Related Work}

\subsection{Parameter-Efficient Fine-Tuning Methods}
Several PEFT methods have been proposed to fine-tune large pre-trained models for downstream tasks. In this section, we introduce several widely used PEFT approaches.

\textbf{Adapters:} Adapters are small, trainable modules typically inserted into pre-trained models, with the original model parameters kept frozen. These modules usually follow a bottleneck architecture that includes a layer normalization (LN), two fully connected (FC) layers, an activation function, and a residual connection. We follow the insertion form consistent with the work~\cite{adapter1}.

\textbf{BitFit:} Bias-Term Fine-Tuning (BitFit)~\cite{bitfit} is a PEFT method that updates only the bias terms and the task-specific classification layer of the model, leaving the other pre-trained parameters unchanged.

\textbf{LoRA:} LoRA adds low-rank matrices to the transformer attention layers. LoRA simulates parameter changes via low-rank decomposition, enabling fine-tuning of large models using a small number of parameters.

\textbf{AdaLoRA:} AdaLoRA~\cite{AdaLoRA} refers to Adaptive Low-Rank Adaptation, which addresses the limitation of distributing the parameter budget evenly across all weight matrices/layers, as seen in LoRA and other methods. AdaLoRA adaptively allocates the parameter budget to different weight matrices based on their importance score and parameterizes the updates using singular value decomposition.

These PEFT methods either update a small subset of pre-trained parameters or introduce additional parameters into a pre-trained backbone network. Consequently, they still require back-propagation through the backbone model, because the gradient values are needed even for frozen layers to update the parameters of fine-tuned layers, as shown in Fig.~\ref{fig:PEFT_MEFT}~(a), leading to high GPU memory consumption.

\subsection{Existing Work in Dialect Identification}

Fine-tuning pre-trained speech models has become a widely adopted approach, as it effectively leverages the knowledge learned from large-scale datasets during pre-training, resulting in improved performance on resource-limited dialect datasets.

Shaik et al.~\cite{DI-FT-1} combined self-supervised adaptive pre-training with fine-tuning on the pre-trained wav2vec 2.0~\cite{wav2vec2} model, achieving high performance in both language and dialect identification. Garg et al.~\cite{DI-FT-3} fine-tuned the pre-trained Conformer~\cite{Conformer} model for classifying African-American and non-African-American English, improving speech recognition performance. Shen et al.~\cite{DI-FT-2} fine-tuned self-supervised pre-trained speech models as feature extractors of the dialect identification system.

Several studies have explored the application of PEFT methods for DI tasks. Radhakrishnan et al.~\cite{DI-PEFT-1} employed Adapters to fine-tune the pre-trained Whisper~\cite{Whisper} model for Arabic dialect identification, achieving state-of-the-art accuracy on the ADI-17~\cite{ADI-17} dataset. Kamiya et al.~\cite{DI-PEFT-2} applied Adapters to the pre-trained wav2vec 2.0 model for DI and Automatic Speech Recognition (ASR) tasks, leveraging DI to enhance Japanese dialect speech recognition.

Despite the success of these studies, they are all based on vanilla fine-tuning or PEFT methods, therefore, suffer from high resource demands.

\section{Method}
We explore three MEFT methods for resource-efficient fine-tuning of pre-trained models on DI tasks: LST~\cite{LST}, UniPT~\cite{UniPT}, and SHERL~\cite{SHERL}.

\subsection{Ladder Side-Tuning}
Ladder Side-Tuning (LST)~\cite{LST}, as illustrated in Fig.~\ref{fig:PEFT_MEFT} (b), constructs a lightweight, trainable side network that runs alongside the backbone model. Each layer in the side network receives input from the intermediate activations via shortcut connections, forming a parallel ladder structure. The $i^{th}$ side layer takes the output from the previous side layer, denoted as $h_{i-1}^g$, and the downsampled intermediate activations, $h_i^f$, from the $i^{th}$ backbone layer. The activations are combined using a trainable gating mechanism: $\mu_i * h_i^f + (1-\mu_i) * h_{i-1}^g$, where $\mu_i = \text{sigmoid}\left(\frac{\alpha_i}{T}\right)$, $\alpha_i$ is a learnable scalar initialized to zero, and $T$($= 0.1$) is a constant temperature. 

By introducing direct connections between layers in the side network, LST eliminates the need for backpropagation through the backbone model. Although constructing the side network using transformer blocks outperforms in natural language processing (NLP) and vision-and-language (VL) tasks, we choose bottleneck Adapter blocks to construct the side network in our work. 
This is because the sequence length of speech features is generally longer than that of text tokens~\cite{sequence_length}, and the space complexity of the standard self-attention mechanism is $O(n^2)$~\cite{O2}. 
We opt not to introduce attention mechanisms in the side network of LST to avoid excessive memory requirements for longer sequences.

\subsection{Universal Parallel Tuning}
Universal Parallel Tuning (UniPT)~\cite{UniPT}, illustrated in Fig.~\ref{fig:PEFT_MEFT} (c), aims to address the potential semantic inconsistencies introduced by the static gating mechanism in LST, which combines the outputs of each backbone layer with its corresponding side layer. UniPT leverages the final layer's superior representation and transfer capabilities in the pre-trained network to overcome this limitation, using its layer feature $F_N$ as a guide. 
In the interaction layer, the feature from $i^{th}$ backbone layer $F_i$ serves as the key and value, while the $F_N$ acts as the query. Feature extraction is performed via a cross-attention operation: \(
F_i' = \left( \lVert \text{ReLU} (F_N F_i^{\top}) \rVert_1 + I \right) F_i.
\)
In the confidence layer, $F_N$ guides the aggregation of features through confidence-based calculations. This interaction ensures semantic coherence between layers.

\subsection{SHERL}
SHERL~\cite{SHERL}, illustrated in Fig.~\ref{fig:PEFT_MEFT} (d), addresses two main issues in MEFT methods: feature redundancy during cross-layer aggregation, which may dilute small but meaningful features, and misalignment between the output projections of the backbone network layers and the original input-output projections.
SHERL first computes the cosine similarity between the features of the first \(N-2\) layers and sums the values along the embedding dimension to obtain the redundancy rate of each feature. It then uses the output of the \((N-1)^{th}\) layer as guidance, with the feature set of shallow layers as the key and value. Feature extraction is performed through cross-attention, while redundancy is reduced based on the calculated redundancy rate.
After the early aggregation, as shown in Fig.~\ref{fig:PEFT_MEFT}, SHERL combines the aggregated early features with those from the \((N-1)^{th}\) layer through gate addition, then passes the result through the \(N^{th}\) layer to produce the final output.

%\newpage

\begin{table*}[t]
\caption{Comparison of different methods on six Mandarin subdialects from the KeSpeech~\cite{kespeech} dataset. We report the overall identification accuracy on the test set, GPU memory usage, and training time per epoch in minutes. The Adapter, LoRA, and AdaLoRA methods are evaluated with hidden dimension (dim), rank (r), or initial rank (init\_r) of \{64, 128, 256\}. The LST, UniPT, and SHERL methods are evaluated with reduction factors ($RF$) of \{2, 4, 8\}. The training time is averaged over all training epochs.} 
\label{table:results}
\centering
% \vspace{-3pt}
\fontsize{14}{15}\selectfont 
\scalebox{0.75}{%
\begin{tabular}{lcccc}
\toprule
\textbf{Method}& \textbf{Trainable Ratio (\%)}  & \textbf{GPU Mem. (GiB) ($\downarrow$)} & \textbf{Time/epoch (min) ($\downarrow$)} & \textbf{Accuracy (\%) ($\uparrow$)}  \\ \midrule
Kaldi-xvector~\cite{kespeech} & —— & —— & ——& 56.34 \\
ResNet-34~\cite{kespeech} & —— & —— & ——& 61.13 \\ 
ECAPA-TDNN~\cite{kespeech} & —— & ——  & —— & 60.77 \\ \midrule \midrule
\textcolor{gray!70}{Head Tuning} & \textcolor{gray!70}{0.22} & \textcolor{gray!70}{7.06} & \textcolor{gray!70}{5.68} & \textcolor{gray!70}{59.33} \\
Vanilla Fine-Tuning & 96.48 & 76.14 & 16.58 & 78.67 \\
Adapter$_{dim=64}$ & 2.88 & 64.21 & 16.21& 76.26 \\
Adapter$_{dim=128}$ & 5.34 & 65.34 & 16.33& 75.83 \\
Adapter$_{dim=256}$ & 9.91 & 67.34 & 16.50& 75.54 \\
LoRA$_{r=64}$ & 2.82 & 53.14 & 15.92& 78.06 \\
LoRA$_{r=128}$ & 5.28 & 53.76&15.99& 78.37 \\
LoRA$_{r=256}$ & 9.85 & 54.95&16.09& \colorbox{green!40}{\strut 78.77}\\
AdaLoRA$_{init\_r=64}$ &2.82 & 52.87& 18.31& 76.86\\
AdaLoRA$_{init\_r=128}$ & 5.29 & 53.44 & 18.35& 77.71 \\
AdaLoRA$_{init\_r=256}$ & 9.86 & 56.87 & 18.49& 77.80 \\
BitFit & 0.33 & 47.33 & 15.05 & 68.83 \\ 
\midrule
LST$_{RF=2}$& 7.09 & 28.37&9.92& 77.37 \\
LST$_{RF=4}$& 3.80 & 22.88&9.17& 77.18 \\
LST$_{RF=8}$& 2.06 & \colorbox{green!40}{\strut 20.39}& \colorbox{green!20}{\strut 8.58}& \colorbox{green!30}{\strut 78.00}\\ 
UniPT$_{RF=2}$& 5.31 & 43.32&17.40& 77.47 \\
UniPT$_{RF=4}$& 2.85 &36.76&16.25& 77.58 \\
UniPT$_{RF=8}$& 1.57 & 34.26&15.51& 76.43 \\
SHERL$_{RF=2}$&7.92 & 45.08&10.78& 75.98 \\
SHERL$_{RF=4}$& 3.30 & 31.64&9.03& 75.43 \\
SHERL$_{RF=8}$& 1.54 & \colorbox{green!20}{\strut 21.54}&\colorbox{green!40}{\strut 8.02}& 75.88 \\ \bottomrule
\end{tabular}
}
\end{table*}

\section{Experiments}

\subsection{Model}
Our experiments use the multilingual Whisper-small~\cite{Whisper}, a transformer-based, general-purpose speech recognition model pre-trained on 680,000 hours of multilingual and multitask supervised data, demonstrating strong speech representation capabilities. 
The embedding dimension of the model is 768.
The input audio is 30 seconds long, converted into a log-Mel spectrogram with a sequence length 1500. If the audio duration is shorter than 30 seconds, it is padded with zeros; if it exceeds 30 seconds, it is truncated to ensure a consistent input length.

For our experiments, we use only the encoder portion of the Whisper model. At the end of the encoder, we append a classification head. The output from the final encoder layer is passed through a projection layer, which maps the feature dimensions to the appropriate space for classification. A pooling operation is then applied to average the features across all time steps of each sample, resulting in a fixed-length pooled representation. Finally, this pooled output is fed into a fully connected classifier to generate the final classification logits. In our configuration, the projection layer has a dimension of 256, and the model has 88 million parameters.

\subsection{Dataset}
% We evaluate our model on six Mandarin subdialects from the KeSpeech~\cite{kespeech} dataset, which aligns with the data selection for dialect identification in the KeSpeech baseline experiments~\cite{kespeech}. The training set consists of 309.6 hours of audio data, while the test set contains 22.1 hours.
We evaluate our model on KeSpeech\cite{kespeech}, which is a large-scale open-source speech dataset designed to advance subdialect identification, featuring 1,542 hours of audio from 27,237 speakers across 34 cities in China. It covers standard Mandarin and 8 subdialects, with rich labeling (transcriptions, speaker identity, subdialect) that enables effective training and evaluation of dialect-related tasks. The inclusion of parallel recordings in both standard Mandarin and regional subdialects supports novel applications like subdialect style conversion, while the extensive speaker diversity and two-phase recordings make it ideal for robust, time-aware dialect modeling.

\subsection{Training Settings}

We use the Adam optimizer for all methods in our experiments. We explore learning rates of \{$1 \times 10^{-3}$, $5 \times 10^{-4}$, $1 \times 10^{-4}$\} and select the optimal value based on evaluation performance. Any extra parameters introduced into the pre-trained model are randomly initialized. All methods are fine-tuned for 10 epochs in each experimental setup until convergence. Training is performed on a single NVIDIA H100 94GB GPU with a batch size of 128 and enables mixed precision training~\cite{mixed_precision}. We do not use gradient checkpointing~\cite{gradient_checkpointing} or quantization~\cite{quantization}. No additional regularization techniques are applied. 

For Adapter, LoRA, and AdaLoRA, we experiment with hidden dimensions, ranks, or initial ranks from \{64, 128, 256\}. For LST, UniPT, and SHERL, we test reduction factors ($RF$) of \{2, 4, 8\}, which control the embedding dimensions of the downsampling layers, upsampling layers, and the side network. The hidden dimension of the Adapter module in LST is fixed at 256. In vanilla fine-tuning, the two convolutional layers at the bottom of the Whisper model—responsible for processing input representations—and the position embedding layer are frozen. In addition to vanilla fine-tuning, PEFT, and our approaches, we also evaluate head tuning, where only the classification head is updated. For all methods, the classification head is trained during fine-tuning.

\subsection{Experimental Results}

\textbf{Accuracy Performance.} The experimental results are shown in Table~\ref{table:results}. Among PEFT methods, Adapter, LoRA, and AdaLoRA achieve performance comparable to vanilla fine-tuning, significantly outperforming the baseline provided by KeSpeech~\cite{kespeech}, which trains the entire network. 
% Notably, LoRA with \(r=256\) achievs the best performance, surpassing vanilla fine-tuning. 
BitFit, however, performed poorly. Our proposed approaches demonstrated similar performance to PEFT methods. Among them, LST with $RF = 8$ achieves an accuracy of 78.00\%, outperforming AdaLoRA and Adapter, maintaining accuracy within 0.67\% of vanilla fine-tuning and 0.77\% of LoRA.

\textbf{GPU Memory Usage.} As shown in Table~\ref{table:results}, PEFT methods can reduce GPU memory usage compared to vanilla fine-tuning. Among PEFT methods, Adapter reduces memory usage by about 11\% to 16\%, while LoRA and AdaLoRA reduce memory usage by approximately 28\% to 30\% compared to vanilla fine-tuning. BitFit achieves the most significant memory reduction at 37.83\%, though its performance was suboptimal. MEFT methods demonstrated a further reduction in GPU memory usage compared to PEFT methods. When \(RF=8\), LST, UniPT, and SHERL reduce GPU memory usage by 73.25\%, 55.00\%, and 71.71\%, respectively, compared to vanilla fine-tuning. LST achieves the best performance among MEFT methods, with the lowest memory usage, 2.3$\times$ lower than the PEFT method with the lowest memory usage.

\textbf{Training Speed.} The average training time per epoch recorded in Table~\ref{table:results} shows that PEFT methods offer limited improvements in training speed. In contrast, among MEFT methods, LST and SHERL significantly enhance training speed. When \(RF=8\), LST and SHERL accelerate training by 1.9$\times$ and 2.1$\times$, respectively. Notably, we observe that the convergence speed of MEFT methods is comparable to PEFT methods.

\section{Discussion}
In our experiments, we found that among three MEFT approaches the training time of UniPT is significantly longer than that of LST and SHERL, with its GPU memory usage notably higher than LST and exceeding SHERL when \(RF = \{8, 16\}\). We attribute this to the cross-attention operations introduced at each layer. This operation's time and space complexity is \(O(n^2)\), which increases computational complexity and GPU memory usage as the sequence length grows. SHERL introduces this operation only once, while LST, in our setup, does not include attention operations. This reveals that for speech models, it is crucial to minimize the introduction of such operations when fine-tuning with MEFT methods to maintain resource efficiency.

In addition to the MEFT methods implemented in our work, we also tested the memory-efficient zeroth-order optimizer (MeZO)~\cite{MeZO}, which estimates gradients using only forward passes and parameter perturbations for optimization. However, we found that MeZO failed to converge, possibly due to the higher effective rank of the Hessian matrix of the loss~\cite{MeZO} when fine-tuning the pre-trained speech model for the DI task.

Furthermore, for all MEFT methods, besides using the output from the final encoder layer of Whisper model, we experimented with the weighted sum of the outputs from all encoder layers, which performed worse than using only the output from the final encoder layer.

Our future work will focus on further reducing MEFT’s resource consumption by integrating advanced memory‐saving techniques (e.g., quantization) and conducting ablation studies to identify and minimize the number of layers requiring fine‐tuning. We also plan to extend our work to utilize MEFT methods in broader speech-related tasks.

\section{Conclusion}
% This paper introduces a resource-efficient dialect identification (DI) approach based on backbone-independent Memory-Efficient Fine-Tuning (MEFT) methods, marking its first application in this domain. 

By leveraging a general-purpose pre-trained speech processing model and implementing MEFT approaches, we demonstrated competitive performance on the DI task using the KeSpeech dataset for six Mandarin subdialects. Our proposed approach significantly reduces GPU memory usage by up to 73.25\% and accelerates training iteration speed by up to 2.1$\times$ compared to the vanilla fine-tuning while maintaining accuracy within 0.67\% of vanilla fine-tuning and 0.77\% of resource-hungry PEFT approaches like LoRA. These results highlight that MEFT approaches, which have recently been adopted for natural language processing, have the potential to be efficiently used in speech-related tasks.

% \begin{thebibliography}{1}

% \bibitem{1}
% G.~Eason, B.~Noble, and I.~N.~Sneddon, ``On certain integrals of
% Lipschitz-Hankel type involving products of Bessel functions,''
% \emph{Phil. Trans. Roy. Soc. London,} vol. A247, pp. 529-551, April
% 1955.

% \bibitem{2}
% J.~Clerk~Maxwell, \emph{A Treatise on Electricity and Magnetism,}
% 3$^{\rm rd}$ ed., vol. 2. Oxford: Clarendon, 1892, pp.68-73.

% \bibitem{3}
% I.~S.~Jacobs and C.~P.~Bean, ``Fine particles, thin films and exchange
% anisotropy,'' in \emph{Magnetism,} vol. III, G.T. Rado and H. Suhl,
% Eds. New York: Academic, 1963, pp. 271-350.

% \bibitem{4}
% K.~Elissa, ``Title of paper if known,'' unpublished.

% \bibitem{5}
% R.~Nicole, ``Title of paper with only first word capitalized,''
% \emph{J. Name Stand. Abbrev.,} in press.

% \bibitem{6}
% Y.~Yorozu, M.~Hirano, K.~Oka, and Y.~Tagawa, ``Electron spectroscopy
% studies on magneto-optical media and plastic substrate interface,''
% \emph{APSIPA Transl. J. Magn. Japan,} vol. 2, pp. 740-741, August 1987
% [\emph{Digests 9$^{\rm th}$ Annual Conf. Magnetics Japan,} p. 301,
% 1982].

% \bibitem{7}
% M.~Young, \emph{The Technical Writer's Handbook.} Mill Valley, CA:
% University Science, 1989.

% \end{thebibliography}

\printbibliography

\end{document}